\documentclass[review,10pt]{elsarticle}

\usepackage{times}
\usepackage{latexsym}
\usepackage{url}

\usepackage{xspace}
\usepackage{graphicx}
\usepackage{multirow}
\usepackage{makecell}
\usepackage{lscape}
\usepackage{bm}
\usepackage{algorithm}
\usepackage{algpseudocode}
\usepackage{framed}
\usepackage{color}
\usepackage{xcolor}
\usepackage{booktabs,colortbl}
\usepackage{bbm}
\usepackage{subfigure}
\usepackage{multirow} 
\usepackage{amssymb}
\usepackage{amsmath}
\usepackage{soul}
\usepackage{multirow}
\usepackage{tabularx}

\usepackage{booktabs}
\usepackage{setspace}
\usepackage{utfsym}

\newcommand{\ie}{\emph{i.e.,}\xspace}

\newcommand{\baby}{S\textsc{lt-fai}\xspace}

\begin{document}

\begin{frontmatter}



\title{Unsupervised Sentence Representation Learning with Frequency-induced Adversarial Tuning and Incomplete Sentence Filtering}


\author[label1,label2]{Bing Wang}
\author[label1,label2]{Ximing Li\corref{cor1}}
\author[label1,label2]{Zhiyao Yang}
\author[label3]{Yuanyuan Guan}
\author[label1,label2]{Jiayin Li}
\author[label1,label2]{Shengsheng Wang}
\cortext[cor1]{Corresponding author: liximing86@gmail.com}

\address[label1]{{College of Computer Science and Technology, Jilin University, China},
            }

\address[label2]{{Key Laboratory of Symbolic Computation and Knowledge Engineering of Ministry of Education, Jilin University, China},
            }

\address[label3]{{School of Humanities, Jilin University, China},
            }            




\begin{abstract}

Pre-trained Language Model (PLM) is nowadays the mainstay of Unsupervised Sentence Representation Learning (USRL). However, PLMs are sensitive to the frequency information of words from their pre-training corpora, resulting in anisotropic embedding space, where the embeddings of high-frequency words are clustered but those of low-frequency words disperse sparsely. This anisotropic phenomenon results in two problems of similarity bias and information bias, lowering the quality of sentence embeddings. To solve the problems, we fine-tune PLMs by leveraging the frequency information of words and propose a novel USRL framework, namely Sentence Representation Learning with Frequency-induced Adversarial tuning and Incomplete sentence filtering (\baby). We calculate the word frequencies over the pre-training corpora of PLMs and assign words thresholding frequency labels. With them, (1) we incorporate a similarity discriminator used to distinguish the embeddings of high-frequency and low-frequency words, and adversarially tune the PLM with it, enabling to achieve uniformly frequency-invariant embedding space; and (2) we propose a novel incomplete sentence detection task, where we incorporate an information discriminator to distinguish the embeddings of original sentences and incomplete sentences by randomly masking several low-frequency words, enabling to emphasize the more informative low-frequency words. Our \baby is a flexible and plug-and-play framework, and it can be integrated with existing USRL techniques. We evaluate \baby with various backbones on benchmark datasets. Empirical results indicate that \baby can be superior to the existing USRL baselines. Our code is released in \url{https://github.com/wangbing1416/SLT-FAI}.

\end{abstract}




\end{frontmatter}


\section{Introduction}

\textbf{S}entence \textbf{R}epresentation \textbf{L}earning (\textbf{SRL}) technically aims to map sentences into fixed-length embeddings with rich semantic and syntactic properties of sentences \cite{kiros2015skip,hill2016learning,conneau2017supervised}. It is a cornerstone task in natural language processing, since it is widely acknowledged as a basic step to a variety of downstream natural language understanding tasks such as topic modeling \cite{dieng2020topic,wang2021layer}, sentiment analysis \cite{peng2020adversarial}, and textual similarity \cite{zhou2022problems}. 


Recently, the community has paid more attention to \textbf{U}nsupervised \textbf{SRL} (\textbf{USRL}) approaches, which can be trained with unlabeled collections of sentences \cite{yan2021consert,gao2021simcse}. Generally, in contrast to supervised approaches, USRL can simultaneously save many manual efforts of collecting labeled data and avoid potential task-specific bias of sentence embeddings caused by certain supervised objectives. Nowadays, the mainstay of USRL is the emerging {P}re-trained {L}anguage {M}odel ({PLM}), which, as the name suggests, was pre-trained over large-scale corpus with self-supervised linguistic objectives. For example, the transformer-based model BERT is pre-trained with the objectives of masked language modeling and next sentence prediction \cite{devlin2019bert}; GPT is trained with an auto-regressive text generation task \cite{radford2019language}.
PLMs can capture high-order and long-range dependency in texts, so as to generate strong word embeddings with contextual information. For each sentence, its embedding can be formed by \textbf{averaging the embeddings of its word tokens}.

\begin{figure*}[t]
    \centering
    \includegraphics[width=1.0\textwidth]{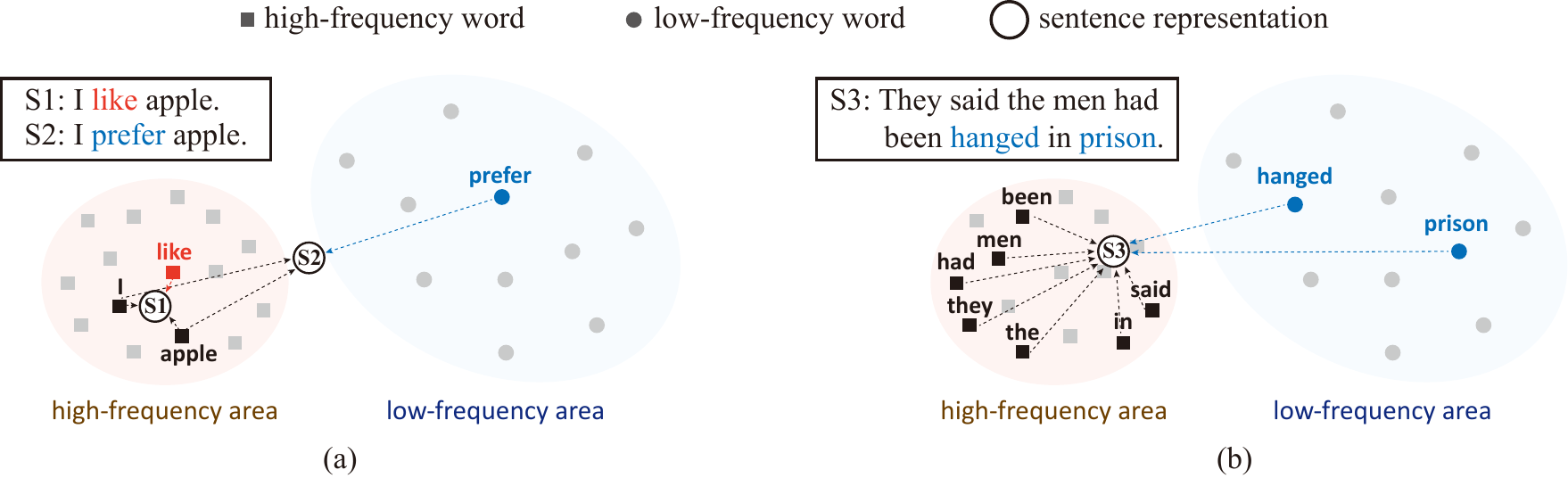}
    \caption{Toy example of an \textit{anisotropic} word embedding space learned by PLMs. Such a phenomenon results in two problems, where we compute each sentence embedding by averaging the embeddings of its word tokens. (a) \textbf{\textit{similarity bias}}: the near-synonyms ``like`` and ``prefer`` are far away in the word embedding space due to their frequency disparity, and the problem is transferred to the sentences S1 and S2 containing them. (b) \textbf{\textit{information bias}}: the embedding of the sentence S3 is far away from the embeddings of ``hanged'' and ``prison'', even they express the underlying semantic information. Best viewed in color.}
    \label{toy}
\end{figure*}


Unfortunately, the previous literature indicate that the word embedding space of PLMs tends to be \textit{anisotropic}, since PLMs are sensitive to the frequency information of words from the pre-training corpus \cite{gao2019representation,ethayarajh2019how,li2020sentence}. The embeddings of high-frequency words are clustered, while those of low-frequency words are dispersed sparsely \cite{yan2021consert}. This phenomenon results in two problems, so as to lower the quality of sentence embeddings of PLMs.

\textit{\textbf{Problem 1: similarity bias.}} Even the semantically relevant words can be far away in the word embedding space due to their frequencies disparity. As an example shown in Fig.\ref{toy}(a), although the words ``like`` and ``prefer`` are semantically related, they lie in the high-frequency area and low-frequency area of the word embedding space, respectively, and are far away from each other. The problem can be directly transferred to the sentence embeddings, especially for relatively shorter sentences.

\textit{\textbf{Problem 2: information bias.}} The sentence embeddings are dominated by high-frequency words, even if the low-frequency words tend to be more informative . As an example shown in Fig.\ref{toy}(b), although the words ``hanged'' and ``prison'' express the underlying semantic information, the sentence embedding lies in the high-frequency area, and is far away from the embeddings of the two words. In some sense, this may result in semantic information loss.

In this paper, we aim to remedy the two problems. Our intuition is since the anisotropic phenomenon is mainly caused by the word frequencies disparity, we can apply the frequency information to fine-tune the outputs of PLMs. Specifically, we calculate the word frequencies over the pre-training corpus of PLMs, and assign words thresholding frequency labels, including high-frequency label and low-frequency label. By applying those frequency labels, we then propose two strategies of \textbf{adversarial tuning} and \textbf{incomplete sentence filtering} to alleviate the similarity bias and information bias, respectively.

\noindent\textbf{\textit{Adversarial tuning for similarity bias.}}
Basically, what we expect is that the resulting word embedding space of PLM is uniformly frequency-invariant. To achieve this, we tune the PLM adversarially by incorporating a \textbf{similarity discriminator} used to distinguish the PLM embeddings of high-frequency words and low-frequency words, while the PLM aims to fool the similarity discriminator to confuse the word embeddings with different frequencies. 


\vspace{0.5pt}
\noindent\textbf{\textit{Incomplete sentence filtering for information bias.}}
We propose a novel incomplete sentence filtering task. For each sentence, we generate its corresponding incomplete sentence by randomly masking several low-frequency words, and then we incorporate an \textbf{information discriminator} used to distinguish the original sentences and incomplete versions. We jointly train the PLM with this task, and it can indirectly emphasize the information contribution of low-frequency words, which are more informative and inspired by the information theory \cite{wilson2010term,li2018exploring}.

Upon those ideas, we propose a novel USRL method, namely \textbf{S}entence \textbf{L}earning \textbf{T}ransfer with \textbf{F}requency-induced \textbf{A}dversarial tuning and \textbf{I}ncomplete sentence filtering (\textbf{\baby}). It is a flexible and plug-and-play framework, so it can be integrated with existing USRL techniques such as the contrastive learning regularization \cite{yan2021consert,gao2021simcse}. In the experiments, we instantiate \baby with various backbones and evaluate on a number of benchmark datasets. Empirical results indicate the effectiveness of \baby, and it can alleviate the anisotropic embedding space to some extent.


In summary, the contributions of this paper can be outlined as the following threefold:

\begin{itemize}

    \item We revisit the anisotropic embedding space of PLMs from the perspective of word frequencies and outline the two problems of similarity bias and information bias for USRL.

    \item To remedy the two problems, we propose a novel USRL framework with frequency-induced adversarial tuning and incomplete sentence filtering, dubbed as \baby.
    
    \item We conduct extensive experiments to evaluate \baby with various backbones on benchmark datasets. Empirical results indicate the effectiveness of \baby.
\end{itemize}





\section{Related Work}

\subsection{Sentence Representation Learning}


Transformer-based PLMs \cite{vaswani2017attention} have received a lot of attention in the natural language processing community. These PLMs \cite{vaswani2017attention} can be divided into two categories: 
(1) text representation with Transformer encoder-only models, pre-trained with the self-supervised objectives such as masked language modeling and next sentence prediction \cite{devlin2019bert,liu2019roberta}; 
and (2) text generation with encoder-decoder or decoder-only modules, pre-trained in an auto-regressive manner \cite{radford2019language,raffel2020exploring}. 
Since PLMs can output strong text representations, fine-tuning them can achieve competitive performance in downstream tasks, ranging from supervised \cite{chalkidis2019large,zhai2022binary}, semi-supervised \cite{chen2020mixtext, li2021semi,cui2022self}, to weakly supervised learning tasks \cite{meng2020text,ouyang2022weakly}.

Recently, PLMs have been employed to prompt SRL, for example, fine-tuning PLMs by decoupling semantics and syntax in sentence embeddings \cite{huang2021disentangling} and leveraging auxiliary word dictionary and definition sentences \cite{tsukagoshi2021defsent}. Despite the success of PLMs in representation learning, \citet{gao2019representation} find that the word embedding space of PLMs tends to be anisotropic, which reveals that the learned word embeddings of PLMs are distributed in a narrow cone. 
Moreover, \citet{ethayarajh2019how,li2020sentence} also argue that the sentence embeddings from PLMs, especially the average contextual embeddings \cite{reimers2019sentence}, suffer from the same anisotropic problem, so as to lower the quality of sentence embeddings. 
To mitigate this issue, some studies propose distribution shift strategies to align PLMs embeddings into an isotropic distribution \cite{li2020sentence,su2021whitening}. 
Additionally, most existing USRL methods fine-tune PLMs by contrastive learning objectives \cite{chen2020a,he2020momentum}, which can indirectly alleviate the anisotropic problem. Contrastive learning is a kind of unsupervised representation learning method in the computer vision community, which can expand embedding spaces and obtain discriminative visual representations. Inspired by this characteristic, some cutting-edge methods transfer this technique into USRL.
For example, \citet{yan2021consert,gao2021simcse} generate positive sentence pairs by various prevalent data augmentation techniques such as adversarial attacks, cutoff, and model dropout to conduct unsupervised contrastive learning; \citet{chuang2022diffcse} introduce the equivariant contrastive learning method \cite{dangovski2021equivariant}, which is proposed to improve visual representation learning, into USRL; \citet{zhou2022debias,klein2022scd} argue that negative sentences are more important for contrastive learning-based USRL, and concentrate on the construction of negative samples, unbiasedly sampled from learned Gaussian distributions or self-contrasted by a strong perturbation. Instead, \citet{tan2022a} initialize a fixed-length pseudo sequence, which is utilized to improve positive samples in latent semantic spaces.

In contrast to the existing USRL methods, we revisit the problems of similarity bias and information bias, caused by the anisotropic phenomenon, from the word frequency perspective. Accordingly, we calculate the word frequencies over the pre-training corpus of PLMs, and formulate two new frequency-induced objectives to solve the problems.


\subsection{Adversarial Learning}

Adversarial-based strategies are always utilized to boost the robustness of deep learning models. These adversarial methods can also be roughly divided into (1) adversarial attacks learn an adaptive noise, in order to confuse deep learning models to make incorrect predictions. With the help of the noises, the vulnerability of the models is mitigated. For instance, 
\citet{yan2021consert} utilize a gradient-based strategy to generate adversarial samples by injecting a
worst-case noise into sentences, and regard the adversarial samples as augmented data items to learn sentence embeddings; \citet{zhang2022improving} design a kind of information bottleneck-based method, which filters out non-robust features, to defend adversarial attacks for text classification.
(2) Adversarial networks \citep{goodfellow2014generative,mirza2014conditional,ganin2015unsupervised} design an adversarial objective to optimize the parameters of the different modules alternately with a MIN-MAX strategy, such as generative adversarial network (GAN) \citep{goodfellow2014generative}. 
This adversarial training paradigm is also adapted to unsupervised domain adaptation \cite{ganin2015unsupervised,li2019transferable,zhou2021an}. They promote the feature extractor to fool the specific domain discriminator, so that the features from the extractor are domain-invariant. Inspired by adversarial networks, we aim to alleviate the similarity bias of PLM-based USRL methods, and we adversarially tune PLMs to achieve uniformly frequency-invariant embedding space.



\section{The Proposed \baby Framework} \label{sec3}

In this section, we introduce the proposed PLM-based USRL framework named \textbf{\baby}. 

\begin{figure*}[t]
    \centering
    \includegraphics[width=1.05\textwidth]{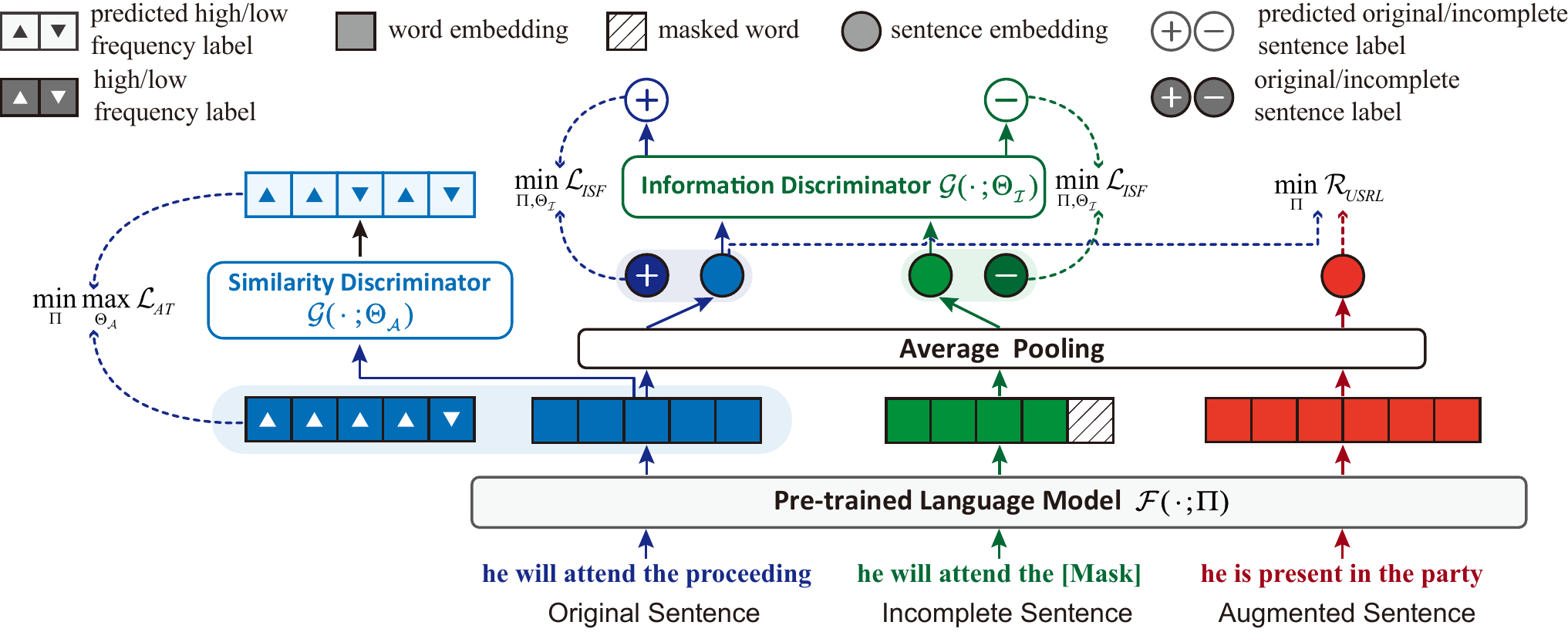}
    \caption{The framework of \baby. We calculate the word frequencies over the pre-training corpora of PLMs, and assign tokens thresholding frequency labels. Given those frequency labels, we propose two strategies of adversarial tuning and incomplete sentence filtering, and instantiate the generic USRL objective by contrastive learning. Best viewed in color.}
    \label{framework}
\end{figure*}

Formally, we are given by a collection of $N$ raw sentences $\Omega = \{\mathbf{s}_i\}_{i=1}^N$, where each raw sentence contains a sequence of $|\mathbf{s}_i|$ word tokens $\mathbf{s}_i=\{w_{ij}\}_{j=1}^{|\mathbf{s}_i|}$. The mainstream of USRL is based on the PLMs, which were pre-trained on large-scale corpora. The PLM $\mathcal{F}(\cdot \:; \Pi)$ can ingest any raw sentence $\mathbf{s}_i$ and output its word embeddings $\mathbf{H}_i = \mathcal{F}(\mathbf{s}_i; \Pi) \in \mathbb{R}^{|\mathbf{s}_i| \times D}$, where $D$ denotes the embedding dimension. It can then form the sentence embedding by the following average pooling:
\begin{equation}
    \label{Eq1}
    \mathbf{h}_i = \frac{1}{|\mathbf{s}_i|} \sum \nolimits _{j=1}^{|\mathbf{s}_i|} \mathbf{H}_{ij}.
\end{equation}
In this work, we propose \baby to fine-tune the PLM $\mathcal{F}(\cdot\:; \Pi)$ over $\Omega$, so as to form stronger sentence embeddings with richer semantics beyond the backbone PLM.  

\subsection{Overview of \baby}

We concentrate on the problems of \textit{similarity bias} and \textit{information bias} caused by the anisotropic phenomenon of PLMs, where the embeddings of high-frequency words are clustered and those of low-frequency words are dispersed sparsely. In \baby, we calculate the word frequencies over the pre-training corpus of PLMs, and assign each token $w_{ij}$ a thresholding frequency label $y_{ij}^\mathcal{A} \in \{0,1\}$, where $0/1$ indicates the high-/low-frequency label.\footnote{We will describe how to tag these labels later.} With those frequency labels, we propose two strategies of (1) \textbf{adversarial tuning} with a similarity discriminator $\mathcal{G}(\cdot\:; \Theta_\mathcal{A})$ and (2) \textbf{incomplete sentence filtering} with an information discriminator $\mathcal{G}(\cdot\:; \Theta_\mathcal{I})$. As a flexible and plug-and-play USRL framework, \baby can be integrated with existing USRL regularization, denoted by $\mathcal{R}_{USRL}$. Upon these components, the overall objective of \baby can be formulated below:
\begin{equation}
    \label{Eq2}
    \mathop{\rm{min}} \limits_{\Pi, \Theta_\mathcal{I}} \mathop{\rm{max}} \limits_{\Theta_\mathcal{A}} \: \alpha \mathcal{L}_{AT}(\Pi, \Theta_\mathcal{A}) 
     + \beta \mathcal{L}_{ISF} (\Pi, \Theta_\mathcal{I}) + \mathcal{R}_{USRL}(\Pi),
\end{equation}
where $\alpha$ and $\beta$ are controllable hyper-parameters. 

For clarity, we show the overall framework of \baby in Fig.\ref{framework}. More details will be introduced in the following subsections.

\subsection{Adversarial Tuning}

To resolve \textit{Problem 1:  similarity bias}, we expect that the resulting word embedding space of PLM is uniformly frequency-invariant. Inspired by \citet{ganin2015unsupervised,li2019transferable,zhou2021an}, we incorporate the similarity discriminator $\mathcal{G}(\cdot\:; \Theta_\mathcal{A})$ to predict the frequency label $p_{ij}^\mathcal{A} = \mathcal{G}(\mathbf{H}_{ij}; \Theta_\mathcal{A})$ of each token from each sentence, and while the PLM aims to fool this discriminator to confuse the word
embeddings with different frequency labels. Accordingly, we propose an adversarial tuning objective with the following \textit{MIN-MAX} formula:
\begin{equation}
    \label{Eq3}
    \mathop{\rm{min}} \limits_\Pi \mathop{\rm{max}} \limits_{\Theta_\mathcal{A}} \mathcal{L}_{AT}(\Pi, \Theta_\mathcal{A}),
\end{equation}
\begin{equation}
    \label{Eq4}
    \mathcal{L}_{AT} (\Pi, \Theta_\mathcal{A}) = 
    \frac{1}{N} \sum \nolimits_{i=1}^N \frac{1}{\vert \mathbf{s}_i \vert} \sum \nolimits_{j \in \mathbf{s}_i} \ell_{CE}(p_{ij}^\mathcal{A}, y_{ij}^\mathcal{A}),
\end{equation}
where $\ell_{CE}(\cdot, \cdot)$ is the cross-entropy loss.




To efficiently and stably implement the adversarial tuning objective, we adopt the gradient reversal layer (GRL) \cite{ganin2015unsupervised}, which can reverse the gradient $\frac{\partial \mathcal{L}_{AT}}{\partial \Pi}$ into $ - \frac{\partial \mathcal{L}_{AT}}{\partial \Pi}$ during the backpropagation process.

\subsection{Incomplete Sentence Filtering}

To resolve \textit{Problem 2: information bias}, we expect the low-frequency words can contribute more to the sentence embeddings. This is inspired by the information theory \cite{wilson2010term,li2018exploring}, where the high-frequency words such as stopwords contain scarce information, but the low-frequency words tend to be more information-rich. 

For each sentence $\mathbf{s}_i$, we generate its corresponding incomplete version $\mathbf{\widehat s}_i$ by randomly masking its low-frequency words at a sampling ratio of $\epsilon$. The low-frequency words contribute more, if the original sentence $\mathbf{s}_i$ and its incomplete version $\mathbf{\widehat s}_i$ are easier to distinguish. Accordingly, we incorporate the information discriminator $\mathcal{G}(\cdot\:; \Theta_\mathcal{I})$ to identify them, and formulate the following objective of incomplete sentence filtering:
\begin{equation}
    \label{Eq5-1}
    \mathop{\rm{min}} \limits_{\Pi,\Theta_\mathcal{I}} \: \mathcal{L}_{ISF} (\Pi, \Theta_\mathcal{I}),
\end{equation}
\begin{equation}
    \label{Eq5}
    \mathcal{L}_{ISF} (\Pi, \Theta_\mathcal{I}) = \frac{1}{N} \sum \nolimits _{i=1}^N \ell_{CE}(\widehat p_i^\mathcal{I}, y_i^\mathcal{I})
     + \ell_{CE}(p_i^\mathcal{I}, y_i^\mathcal{I}),
\end{equation}
\begin{equation}
    \label{Eq6}
    \widehat p_i^\mathcal{I} = \mathcal{G}(\mathbf{\widehat h}_i; \Theta_\mathcal{I}), \quad p_i^\mathcal{I} = \mathcal{G}(\mathbf{h}_i; \Theta_\mathcal{I})
\end{equation}
where $y_i^\mathcal{I} = \{0,1\}$, and $0$ and $1$ indicate the original sentence and incomplete sentence, respectively.

\subsection{Instantiated USRL Regularization}

To our knowledge, most cutting-edge USRL methods are based on contrastive learning \cite{yan2021consert,gao2021simcse}, which can also alleviate the anisotropic problem indirectly.
Accordingly, we instantiate $\mathcal{L}_{USRL}$ with the contrastive learning regularization.
For each sentence $\mathbf{s}_i$, we generate two augmented versions $\mathbf{s}_i^a$ and $\mathbf{s}_i^b$. We feed them into the PLM to achieve their word embeddings $\mathbf{H}_i^a = \mathcal{F}(\mathbf{s}_i^a; \Pi)$ and $\mathbf{H}_i^b = \mathcal{F}(\mathbf{s}_i^b; \Pi)$, and further form their sentence embeddings $\mathbf{h}_i^a$ and $\mathbf{h}_i^b$ by Eq.~\eqref{Eq1}. The contrastive learning regularization aims to pull the embeddings of augmented versions from one sentence closer and pull the embeddings from different sentences farther in a mini-batch. Specifically, we instantiate the regularization $\mathcal{R}_{USRL}(\Pi)$ as follows:
\begin{equation}
    \label{Eq7}
    \mathop{\rm{min}} \limits_{\Pi} \:  \mathcal{R}_{USRL}(\Pi),
\end{equation}
\begin{equation}
    \label{Eq8}
    \mathcal{R}_{USRL}(\Pi) = 
    - \frac{1}{B}  \sum\limits _{i \in \Omega} \log \frac{\exp(\text{sim}(\mathbf{h}_i^a, \mathbf{h}_i^b) / \tau )}
    {\sum \nolimits _{k \in \Omega_{\neg i}} \exp(\text{sim}(\mathbf{h}_i^a, \mathbf{h}_k) / \tau)},
\end{equation}
where $B$ is the mini-batch size; $\Omega$ denotes the mini-batch; $\Omega_{\neg i}$ denotes subset of the augmented sentences excluding the ones of $\mathbf{s}_i$; and $\text{sim}(\cdot, \cdot)$ is a similarity measure.\footnote{Here, we fix the similarity measure to the cosine similarity.}

Additionally, we generate the augmented sentences for $\mathcal{R}_{USRL}(\Pi)$ by applying the strategies of ConSERT (\ie feature cutoff, token cutoff, token shuffling) \cite{yan2021consert} and SimCSE (\ie model dropout) \cite{gao2021simcse}, respectively. We treat ConSERT and SimCSE as two respective backbones.

\subsection{Implementation of Frequency Label Annotation} \label{sec3.4}

\begin{figure*}[h]
    \centering
    \includegraphics[width=0.7\textwidth]{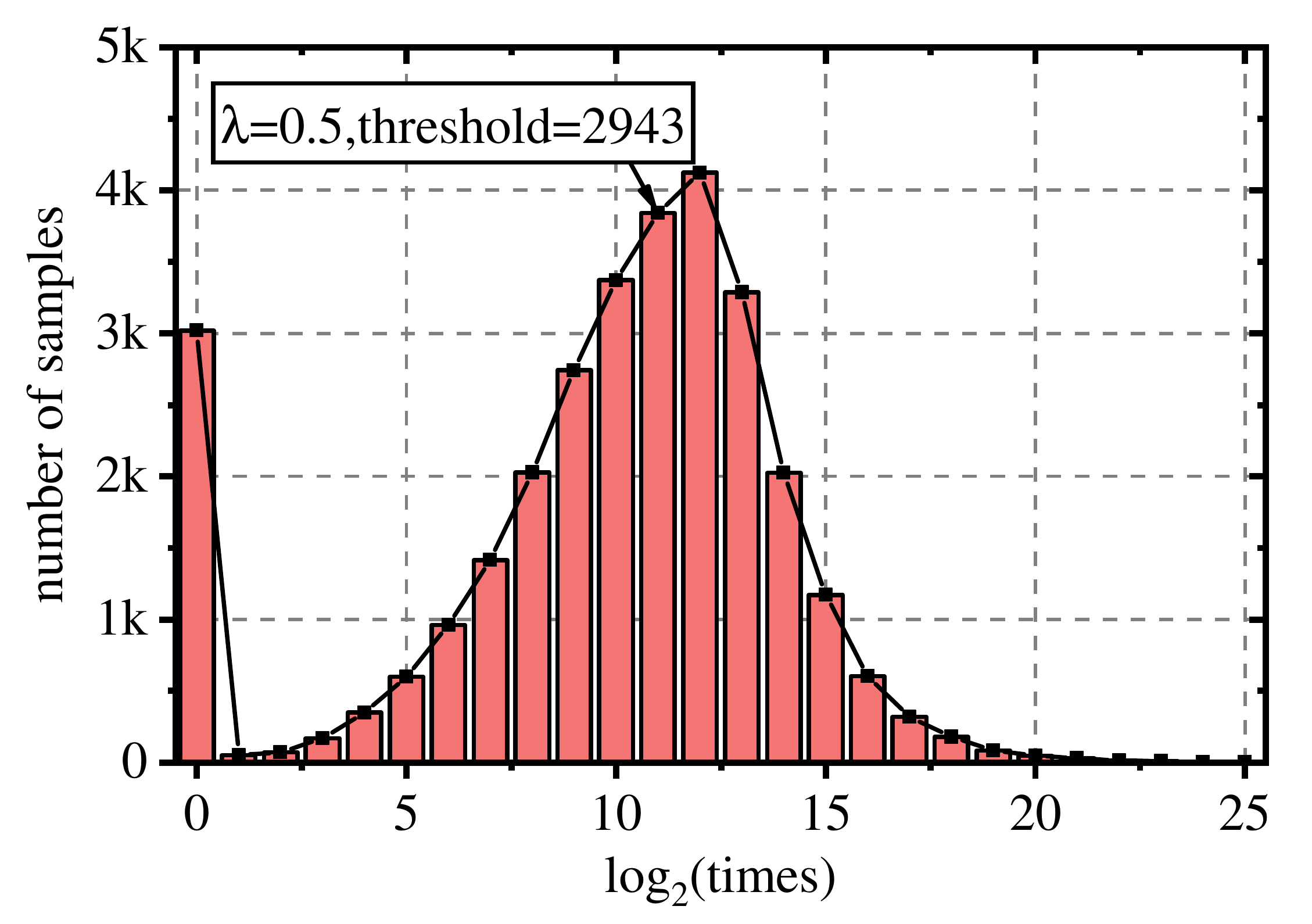}
    \caption{The statistic of frequency labels.}
    \label{label}
\end{figure*}

In this work, we concentrate on BERT \cite{devlin2019bert}, and we collect its per-training corpus \textit{BookCorpus},\footnote{\url{https://huggingface.co/datasets/bookcorpus}} which contains 74M sentences and 1.1B tokens \cite{zhu2015aligning}. We tokenize \textit{BookCorpus} with the \textit{bert-base-uncased}\footnote{\url{https://huggingface.co/bert-base-uncased}} Tokenizer and calculate the frequency of each word, and the specific statistic result is shown in Fig.~\ref{label}. We sort all words according to their frequencies, and assign the frequency label $y^\mathcal{A} = 1$ to $\lambda \in (0, 1)$ low-frequency words and $y^\mathcal{A} = 0$ to the other $1 - \lambda$ words.

\section{Experiments}

\subsection{Experimental Settings}

\vspace{0.5pt}
\noindent \textbf{Datasets.}
In the experiments, we employ two training datasets: (1) 89,192 unlabeled sentences from Semantic Textual Similarity (STS) tasks, including STS 2012 - 2016 (\textit{STS12-STS16}) \cite{agirre2012semeval,agirre2013semeval,agirre2014semeval,agirre2015semeval,agirre2016semeval}, STSBenchmark (\textit{STSb}) \cite{cer2017semeval} and SICKRelatedness (\textit{SICKR}) \cite{marelli2014a}, and (2) randomly sampled 1,000,000 raw sentences from English Wikipedia \cite{gao2021simcse}.

Meanwhile, we evaluate \baby by STS tasks. The test samples are from the test sets of \textit{STS12-STS16}, \textit{STSb}, and \textit{SICKR}. Each test sample consists of pairwise sentences, and an annotated score between 0 and 5 is given to measure their semantic similarity. Following the previous USRL studies \cite{yan2021consert,gao2021simcse}, we apply the Spearman correlation between the annotated scores and the cosine similarity values of sentence embeddings as the evaluation metric.

\vspace{0.5pt}
\noindent \textbf{Baselines.}
Since \baby can be treated as a plug-and-play framework, we reiterate that, to thoroughly evaluate \baby, in the experiment we employ two contrastive learning-based backbones ConSERT\footnote{The code is available at \url{https://github.com/yym6472/ConSERT}} \cite{yan2021consert} and SimCSE\footnote{The code is available at \url{https://github.com/princeton-nlp/SimCSE}} \cite{gao2021simcse}. We also select the two backbones as baseline methods, and comparing with them can directly evaluate the effectiveness of the new objectives of \baby. Additionally, we compare \baby with several other existing USRL methods, including USE \cite{cer2018universal}, CLEAR \cite{wu2020clear}, BERT-flow \cite{li2020sentence}, BERT-whitening \cite{su2021whitening}, IS-BERT \cite{zhang2020an}, CT-BERT \cite{carlsson2021semantic}, SG-OPT \cite{kim2021self}, and SCD-BERT \cite{klein2022scd}.



\vspace{0.5pt}
\noindent \textbf{Implementation details.}
We implement our experiments with the same settings as ConSERT and SimCSE, but the batch size is set to 32, due to the GPU memory limit. We reiterate that we adopt the PLM BERT-base \cite{devlin2019bert}. 
We utilize feed-forward neural networks, which contain two linear layers and a ReLU activation function as the similarity discriminator and information discriminator. And we perform a warm-up stage that the model is trained with $\mathcal{L}_{USRL}$ only, and we fix the warm-up iterations to 0.5 epochs for ConSERT + \baby and 0.1 epochs for SimCSE + \baby. Additionally, the hyper-parameters $\alpha$, $\beta$, and $\lambda$ are fixed to 1.0, 1.0, and 0.5, respectively. The frequency label rate $\lambda$ is fixed as 50\%. In incomplete sentence filtering, the sampling ratio $\epsilon$ is set to 0.2.

\begin{table*}
\centering
\renewcommand\arraystretch{1.25}
\footnotesize
  \caption{Empirical results of \textbf{\baby}. Except for baselines signed with BERT-large, all other comparing methods are based on BERT-base \cite{devlin2019bert}. $^\dag$ indicates the results are rerun by the public codes when the batch size is fixed to 32. The \textbf{bold} and \underline{underlined} scores denote the best results among all comparing methods and ablative versions of \baby, respectively.}
  \label{result}
  \setlength{\tabcolsep}{5pt}{
  \begin{tabular}{m{3.35cm}<{\centering}m{0.7cm}<{\centering}m{0.7cm}<{\centering}m{0.7cm}<{\centering}m{0.7cm}<{\centering}m{0.7cm}<{\centering}m{0.7cm}<{\centering}m{0.7cm}<{\centering}m{0.7cm}<{\centering}}
    \toprule
    Model & \textit{STS12} & \textit{STS13} & \textit{STS14} & \textit{STS15} & \textit{STS16} & \textit{STSb} & \textit{SICKR} & Avg. \\
    \hline
    \specialrule{0em}{0.5pt}{0.5pt}
    \hline
    \multicolumn{9}{c}{\textit{without training}} \\
    GloVe \cite{pennington2014glove} & 55.14 & 70.66 & 59.73 & 68.25 & 63.66 & 58.02 & 53.76 & 61.32 \\
    BERT (\texttt{[CLS]}) & 21.54 & 32.11 & 21.28 & 37.89 & 44.24 & 20.30 & 42.42 & 31.40 \\
    BERT ( pooling) & 30.87 & 59.89 & 47.73 & 60.29 & 63.73 & 47.29 & 58.22 & 52.57 \\
    BERT$_{\text{large}}$ (\texttt{[CLS]}) & 27.44 & 30.76 & 22.59 & 29.98 & 42.74 & 26.75 & 43.44 & 31.96 \\
    BERT$_{\text{large}}$ (average pooling) & 27.67 & 55.79 & 44.49 & 51.67 & 61.88 & 47.00 & 53.85 & 48.91 \\
    \hline
    \specialrule{0em}{0.5pt}{0.5pt}
    \hline
    \multicolumn{9}{c}{\textit{trained with STS dataset (89,192 samples)}} \\
    BERT-flow \cite{li2020sentence} & 63.48 & 72.14 & 68.42 & 73.77 & 75.37 & 70.72 & 63.11 & 69.57 \\
    BERT$_{\text{large}}$-flow \cite{li2020sentence} & 65.20 & 73.39 & 69.42 & 74.92 & 77.63 & 72.26 & 62.50 & 70.76 \\
    \hline
    $^\dag$ConSERT \cite{yan2021consert} & \textbf{65.28} & 77.93 & 68.18 & 78.51 & 74.94 & 72.51 & 66.54 & 71.98 \\
    \rowcolor{lightgray} ConSERT + \textbf{\baby} (\textbf{Ours}) & \underline{65.20} & \underline{\textbf{79.66}} & \underline{\textbf{70.23}} & \underline{\textbf{80.18}} & \underline{\textbf{75.88}} & 73.67 & \underline{\textbf{68.24}} & \underline{\textbf{73.29}} \\
    ConSERT + \baby w/o ISF & 63.67 & 78.71 & 69.66 & 79.77 & 75.82 & \underline{\textbf{74.30}} & 67.64 & 72.80 \\
    ConSERT + \baby w/o AT & 63.69 & 77.55 & 69.70 & 79.81 & 75.86 & 73.92 & 66.85 & 72.48 \\
    \hline
    \specialrule{0em}{0.5pt}{0.5pt}
    \hline
    \multicolumn{9}{c}{\textit{trained with English Wikipedia dataset (1,000,000 samples)}} \\
    USE \cite{cer2018universal} & 64.49 & 67.80 & 64.61 & 76.83 & 73.18 & 74.92 & 76.69 & 71.22 \\
    CLEAR \cite{wu2020clear} & 49.00 & 48.90 & 57.40 & 63.60 & 65.60 & 75.60 & 72.50 & 61.80 \\
    BERT-flow \cite{li2020sentence} & 58.40 & 67.10 & 60.85 & 75.16 & 71.22 & 68.66 & 64.47 & 66.55 \\
    BERT-whitening \cite{su2021whitening} & 57.83 & 66.90 & 60.90 & 75.08 & 71.31 & 68.24 & 63.73 & 66.28 \\
    IS-BERT \cite{zhang2020an} & 56.77 & 69.24 & 61.21 & 75.23 & 70.16 & 69.21 & 64.25 & 66.58 \\
    CT-BERT \cite{carlsson2021semantic} & 61.63 & 76.80 & 68.47 & 77.50 & 76.48 & 74.31 & 69.19 & 72.05 \\
    SG-OPT \cite{kim2021self} & \textbf{77.23} & 68.16 & 66.84 & 80.13 & 71.23 & \textbf{81.56} & \textbf{77.17} & 74.62 \\
    SCD-BERT \cite{klein2022scd} & 66.94 & 78.03 & 69.89 & 78.73 & 76.23 & 76.30 & 73.18 & 74.19 \\
    \hline
    $^\dag$SimCSE \cite{gao2021simcse} & 67.93 & 78.90 & 71.72 & 81.49 & 77.10 & 76.79 & 68.24 & 74.60 \\
    \rowcolor{lightgray} SimCSE + \textbf{\baby} (\textbf{Ours}) & \underline{67.80} & 82.12 & 73.22 & \underline{\textbf{82.42}} & {77.69} & \underline{78.85} & \underline{71.19} & \underline{\textbf{76.18}} \\
    SimCSE + \baby w/o ISF & 67.56 & 81.40 & \underline{\textbf{73.39}} & 81.21 & \underline{\textbf{78.26}} & 77.59 & 70.22 & 75.66 \\
    SimCSE + \baby w/o AT & 66.48 & \underline{\textbf{82.16}} & 73.48 & 81.40 & 77.16 & 76.58 & 70.25 & 75.36 \\
    \bottomrule
  \end{tabular} }
\end{table*}

\subsection{Main Results and Ablation Study}

The Spearman correlation scores of all comparing methods are reported in Table~\ref{result}. 
Overall speaking, it can be clearly seen that our \baby outperforms the baseline methods in most settings, and especially \baby achieves the highest scores on average. Compared with the two backbones, we can observe that the average scores of \baby exceed ConSERT and SimCSE by 1.31 and 1.59, respectively. 
It is worth noting that our \baby is a plug-and-play method, so the empirical results are enough to prove that \baby can effectively improve the performance of the baselines.
Moreover, although ConSERT + \baby is trained on the small STS datasets, it can perform better than most baselines trained with English Wikipedia datasets, which contain much more training sentences.


Turning to the ablative evaluations, we compare among different ablative versions of \baby, also shown in Table \ref{result}. First, \baby can surpass the versions \baby w/o ISF and \baby w/o AT in most settings. For example, the average scores without ISF are reduced by 0.49 and 0.52 based on ConSERT and SimCSE, while the average scores without AT are reduced by 0.81 and 0.82 based on the two backbones. These results directly indicate that the two novel objectives have positive effects on STS tasks. Besides, \baby w/o ISF is superior to \baby w/o AT to some extent. It implies that aligning the two frequency areas of word embeddings can be more significant to alleviate the anisotropic problem. 





\begin{figure}[t]
    \centering
    \includegraphics[scale=0.30]{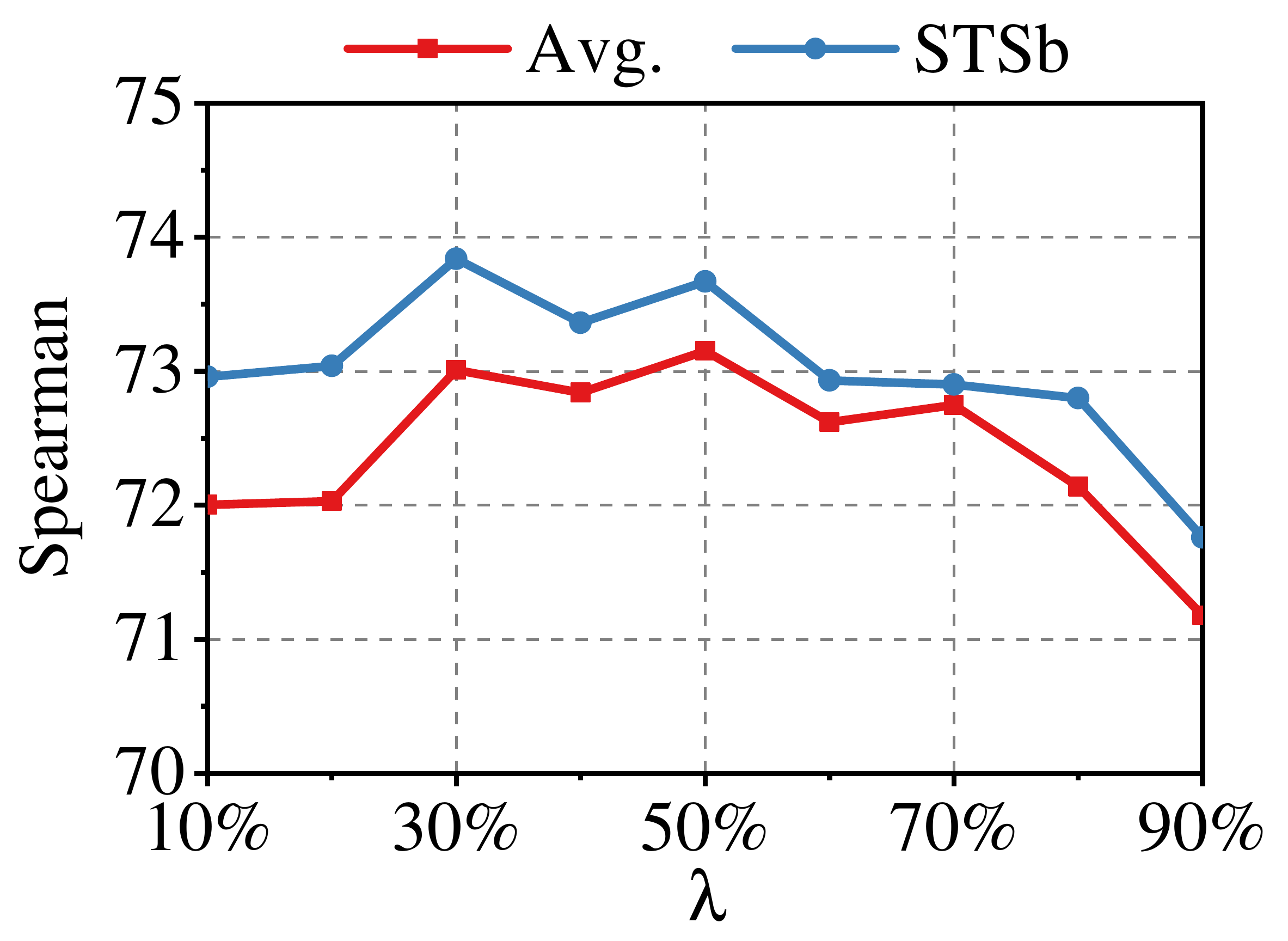}
    \caption{Sensitivity analysis of the frequency label rate $\lambda$.}
    \label{sensitivity}
\end{figure}

\subsection{Parameter Evaluation and Visualization}

We present the results of parameter analysis and visualizations with ConSERT+\baby. Due to the space limit, we omit the results of SimCSE+\baby, which performed similar trends in the early experiments.

\subsubsection{Varying the Frequency Label Rate $\lambda$}

The parameter $\lambda$ is used as a thresholding value to assign frequency labels to words. We analyze its sensitivity by varying its values over the set $\{10\%,30\%,\cdots,90\%\}$. The empirical results are presented in Fig.\ref{sensitivity}. It can be seen that \baby performs better when $\lambda$ is around $30\% \sim 50\%$, and both the best average scores and scores of STSb are happening when $\lambda = 50\%$. This implies an interesting finding that a balanced frequency label assignment is beneficial for \baby. Accordingly, we suggest $\lambda = 50\%$ as the default setting of \baby. 


\begin{figure*}[t]
    \centering
    \includegraphics[scale=0.90]{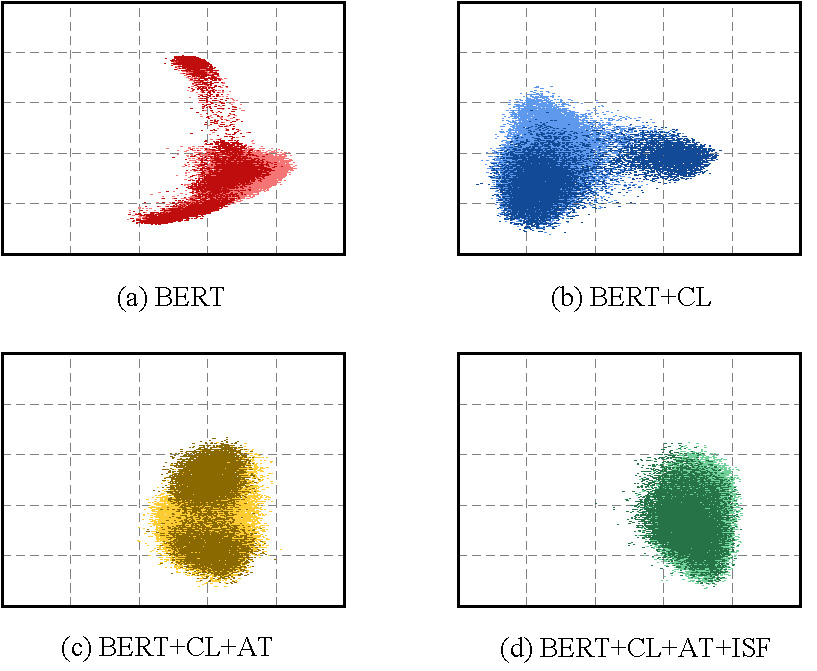}
    \caption{Word embedding visualizations based on the backbone ConSERT. ``CL`` denotes the contrastive learning regularization instantiated by ConSERT. ``AT`` and ``ISF`` denote the adversarial tuning objective $\mathcal{L}_{AT}$ and incomplete sentence filtering objective $\mathcal{L}_{ISF}$, respectively.}
    \label{tokenembedding}
\end{figure*}

\begin{figure*}[t]
    \centering
    \includegraphics[scale=0.90]{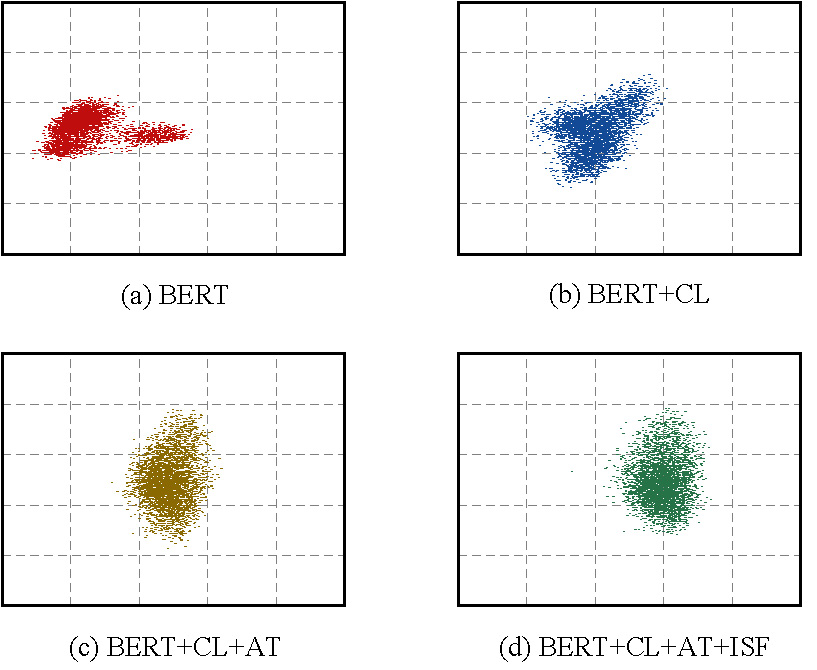}
    \caption{Sentence embedding visualizations based on the backbone ConSERT. ``CL`` denotes the contrastive learning regularization instantiated by ConSERT. ``AT`` and ``ISF`` denote the adversarial tuning objective $\mathcal{L}_{AT}$ and incomplete sentence filtering objective $\mathcal{L}_{ISF}$, respectively.}
    \label{sentenceembedding}
\end{figure*}

\subsubsection{Embedding Visualization}

We randomly draw 10,000 sentences from the STS training datasets, and show a number of 2-dimensional visualizations of the word and sentence embeddings in Figs.\ref{tokenembedding} and \ref{sentenceembedding}. In terms of word embeddings, we can observe that the vanilla BERT results in a clear anisotropic embedding space, and the space gradually becomes isotropic by applying contrastive learning, especially the proposed adversarial tuning and incomplete sentence filtering. With this trend, the sentence embeddings also tend to be isotropic, and it has met our expectations. 


\subsubsection{Self-attention Weight Visualization}

\begin{figure*}[t]
    \centering
    \includegraphics[scale=0.75]{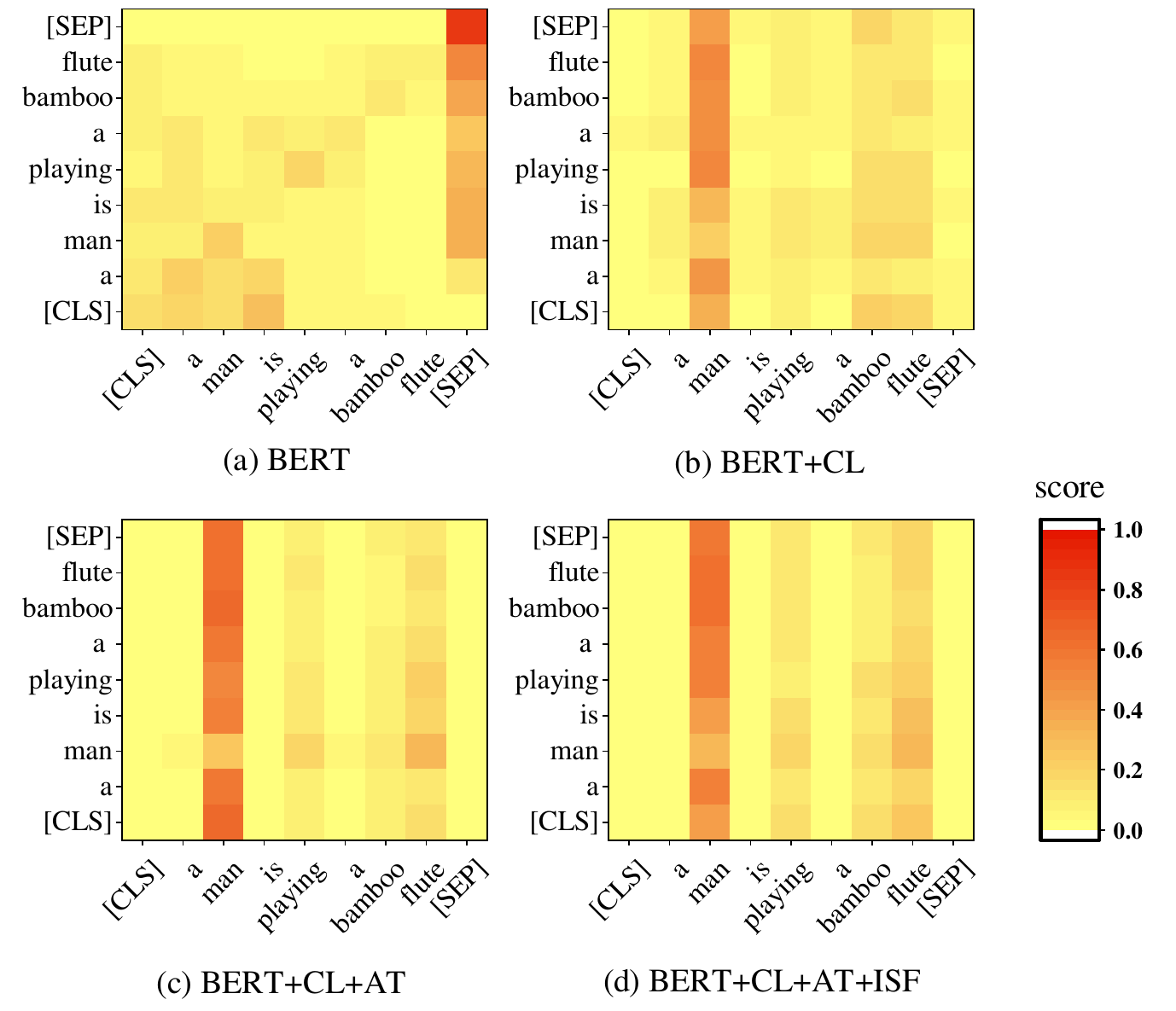}
    \caption{Case study of the self-attention weights among word tokens in the raw sentence ``\texttt{[CLS]} a man is playing a bamboo flute \texttt{[SEP]}''.}
    \label{attention}
\end{figure*}

The incomplete sentence filtering aims to emphasize informative words when forming sentence embeddings. To evaluate its effectiveness, we select a raw sentence ``\textit{a man is playing a bamboo flute}``, and visualize its word self-attention weights of the last Transformer layer in Fig.\ref{attention}. Generally, larger self-attention weights imply more influence to other words, indirectly achieving more influence to the sentence. In terms of vanilla BERT, we can see that the token \texttt{[SEP]} corresponds to larger self-attention weights but it is meaningless. In contrast, the outputs of \baby, especially with incomplete sentence filtering, have larger self-attention weights for the words ``man'', ``playing'', ``bamboo'', and ``flute'', which are obviously more informative, indicating the effectiveness of \baby. This is consistent to the observations in embedding visualization evaluations.

\section{Conclusion and Limitation}

In this paper, we concentrate on the similarity bias and information bias caused by the anisotropic phenomenon of PLMs. To solve them, we propose a flexible and plug-and-play USRL framework \baby which contains adversarial tuning, incomplete sentence filtering, and a baseline USRL regularization. First, we gain inspiration from unsupervised domain adaption and design an adversarial tuning strategy to align high/low-frequency areas in the PLM embedding space. Then, to highlight the importance of low-frequency words in a sentence, we are inspired by information theory, and propose an incomplete sentence filtering objective to achieve this purpose. The empirical performance demonstrates that \baby can exceed most USRL baselines. Additionally, many visualization experiments prove that our \baby can effectively relieve the aforementioned similarity bias and information bias. 

We also discuss the limitations of \baby. 
Despite adversarial tuning and incomplete sentence filtering can consistently improve the performance of our model, and significantly solve the challenges caused by anisotropy, but these objectives also slightly improve the spatial complexity of our framework. For example, incomplete sentence filtering will generate an incomplete version for each sentence, which is worth being solved in the follow-up works.

\section*{Acknowledgments}
We would like to acknowledge support for this project from the National Key R\&D Program of China (No.2021ZD0112501, No.2021ZD0112502), the National Natural Science Foundation of China (No.62276113), and China Postdoctoral Science Foundation (No.2022M721321).




 \bibliographystyle{elsarticle-harv} 
 \bibliography{SLTFAI}





\end{document}